\documentclass[lettersize,journal]{IEEEtran}
\usepackage{amsmath,amsfonts}
\usepackage{algorithmic}
\usepackage{algorithm}
\usepackage{array}
\usepackage[caption=false,font=normalsize,labelfont=sf,textfont=sf]{subfig}
\usepackage{textcomp}
\usepackage{multirow}
\usepackage{stfloats}
\usepackage{url}
\usepackage{xcolor}
\usepackage{graphicx}
\usepackage{cite}
\usepackage{hyperref}
\usepackage{booktabs}
\begin{document}

\title{Multi-Submap Implicit Neural SLAM with Local-to-Global Loop Closure for Large-Scale Scene Reconstruction }

\author{Tianchen Deng, Chongdi Wang, Nailin Wang, Lei Zhao, Ziqi Ma, Tianjun Zhang, Zhe Liu,\\ Danwei Wang,~\IEEEmembership{Life Fellow,~IEEE,} Hesheng Wang,~\IEEEmembership{Senior Member,~IEEE} 
\thanks{Tianchen Deng, Nailin Wang, Lei Zhao, Ziqi Ma, Zhe Liu, Hesheng Wang are are with School of Automation and Intelligent Sensing, Shanghai Jiao Tong university and State Key Laboratory of Avionics Integration and Aviation System-of-Systems Synthesis, Shanghai Key Laboratory of Navigation and Location Based Services, Shanghai 200240, China. Chongdi Wang is with Carnegie Mellon University, Pittsburgh, PA, United States. Danwei Wang is with the School of Electrical and Electronic Engineering, Nanyang Technological University, Singapore. This work was supported by National Key R\&D Program of China (Grant No.2024YFB4708900). It was also supported in part by the Natural Science Foundation of
China under Grant 62225309,U24A20278, 62361166632. The first two authors contribute equal to this paper.  (*corresponding author: wanghesheng@sjtu.edu.cn)
	}}

\markboth{Journal of \LaTeX\ Class Files,~Vol.~14, No.~8, August~2021}%
{Shell \MakeLowercase{\textit{et al.}}: A Sample Article Using IEEEtran.cls for IEEE Journals}


\maketitle
\vspace{-0.4cm}
\begin{abstract}
Neural Radiance Fields (NeRF)-based SLAM has demonstrated impressive results in small-scale scene reconstruction, yet scaling these methods to extensive, complex environments remains challenging due to catastrophic forgetting and accumulated trajectory drift. This paper presents a robust, large-scale neural SLAM system featuring a multi-submap architecture and a dual-tier loop closure mechanism. Specifically, we propose a progressive mapping strategy that dynamically allocates neural submaps to maintain high-fidelity representations without memory explosion. For robust pose estimation, an optical-flow-based tracking module is integrated to handle aggressive motions. To address global consistency, we introduce a local-to-global loop closure framework leveraging the foundation model for high-performance global descriptor extraction, significantly enhancing relocalization accuracy under varying viewpoints. Furthermore, an inter-submap online distillation algorithm is designed during back-end optimization to enforce geometric and appearance consistency across overlapping submap boundaries. To validate the system, we developed a customized handheld mechatronic platform and conducted extensive evaluations on both public benchmarks and our large-scale indoor-outdoor datasets. Experimental results, including direct deployment on an onboard computing unit, demonstrate that our approach outperforms state-of-the-art neural SLAM methods in reconstruction quality and localization robustness, providing a scalable solution for real-world robotic perception and digital twinning. We will release the code publicly on \href{https://github.com/dtc111111/MSN-SLAM}{https://github.com/dtc111111/MSN-SLAM} .
\end{abstract}


\section{Introduction}
Visual Simultaneous Localization and Mapping (SLAM) has been a fundamental problem in robotics, serving as a cornerstone for autonomous navigation and interactive perception~\cite{deng2025best3dscenerepresentation, 10571783, 11126165}. Traditional methods, such as ORB-SLAM \cite{orbslam} and VINS-Mono \cite{vins}, have achieved remarkable success in real-time camera tracking and sparse map construction. However, the sparse point clouds generated by these systems often lack the dense geometric information necessary for advanced robotic tasks. Attention then shifts toward dense scene reconstruction, which aims to provide a continuous and complete geometry for robotic interaction. Early dense methods, as exemplified by DTAM \cite{dtam} and KinectFusion \cite{kinectfusion}, laid the foundation by performing dense regularized reconstruction or Volumetric TSDF (Truncated Signed Distance Function) mapping. However, these approaches demand extremely high memory consumption and suffer from low accuracy during real-time operation.
Recently, with the proposal of Neural Radiance Fields (NeRF), many works focus on combining implicit scene representation with SLAM systems, such as iMAP \cite{imap}, NICE-SLAM \cite{niceslam}, Co-SLAM \cite{coslam} and ESLAM~\cite{eslam}. These methods leverage neural networks as continuous priors to represent scene geometry.

However, these neural implicit methods inherit their own set of challenges: they often rely on a fixed-capacity global MLP or grid, which leads to catastrophic forgetting or memory explosion when the robot transitions from a single room to large-scale, multi-room environments. Second, most neural SLAMs lack robust global loop closure and consistent backend optimization, resulting in significant accumulation of errors and pose drift during extended trajectories.
To this end, we propose a robust Multi-Submap Neural SLAM system designed for large-scale scene reconstruction. Our core contributions are centered around the following four pillars:
First, we present a unified large-scale neural SLAM framework that systematically integrates progressive multi-submap mapping, local-to-global optimization, and online inter-submap distillation to enable scalable and consistent neural scene reconstruction. Second, to mitigate cumulative drift, we design a hierarchical loop closure mechanism. We integrate the SALAD descriptor, which leverages foundation-model-based features, to provide highly discriminative global descriptors. This enables our system to perform robust Local and Global Loop Closures, ensuring accurate relocalization even under drastic viewpoint changes that traditional descriptors fail to handle.
Third, for global consistency across the submap architecture, we introduce a novel online distillation algorithm within the backend optimization. This method enforces geometric and photometric coherence between overlapping submaps by distilling spatial knowledge in real-time. It effectively eliminates "seam" artifacts at submap boundaries and maintains a globally consistent dense map.
Furthermore, bridging the gap between algorithm and hardware, we constructed a handheld mechatronic platform equipped with synchronized sensors. Our system incorporates an optical-flow-based tracking module for robust pose estimation. The entire pipeline is directly validated on onboard computing units, demonstrating SOTA performance in both localization and reconstruction for real-world, large-scale scenarios.

Overall, our contributions are shown as follows:
\begin{itemize}
\item \textbf{Progressive Multi-Submap Representation:} A dynamic submap management strategy that enables scalable neural reconstruction for large-scale scenes while effectively preventing memory explosion and catastrophic forgetting.
\item \textbf{Local-to-Global Loop Closure:} A hierarchical local-to-global loop closure mechanism utilizing foundation-model-based visual descriptors to achieve robust relocalization and global consistency under challenging viewpoint variations.

\item \textbf{Inter-Submap Online Distillation:} A novel backend optimization algorithm that enforces geometric and photometric coherence across overlapping submaps through real-time knowledge distillation, ensuring a seamless global map.

\item \textbf{System Integration and Onboard Validation:} A tightly-coupled mechatronic sensing platform featuring sub-millisecond hardware-level synchronization between LiDAR and cameras, ensuring data consistency for high-fidelity mapping. The development of the handheld mechatronic platform, validated via real-time experiments on embedded computing units in complex environments.
\end{itemize}

\begin{figure*}[ht]
    \centering
    
    \includegraphics[width=0.9\linewidth]{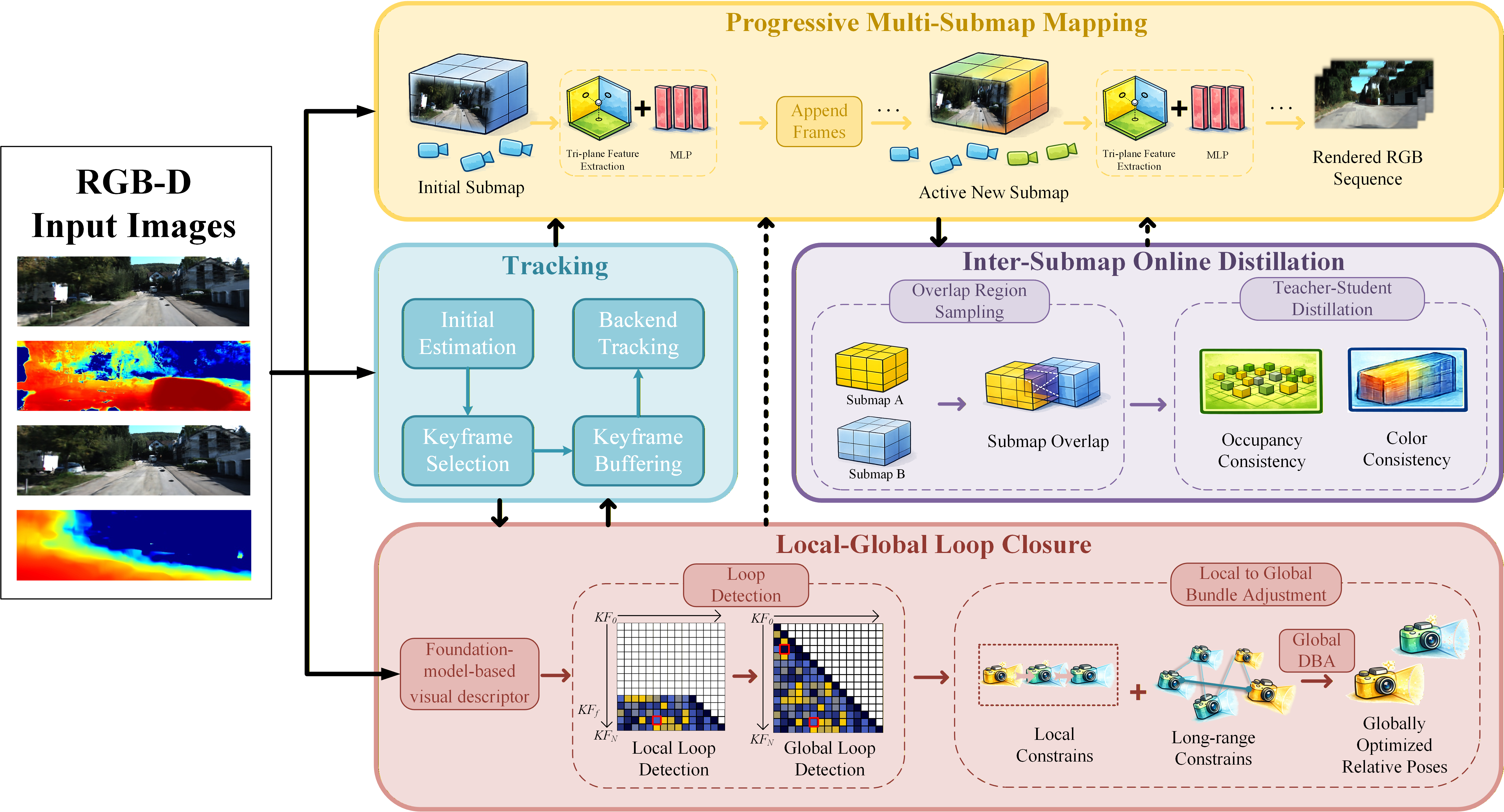}
\vspace{-0.4cm}
    \caption{Overview of the proposed system pipeline. The framework consists of three core components: (a) Progressive multi-submap representation, which partitions the scene into local submaps $S_i$ using a hybrid tri-plane and MLP structure to handle large-scale environments; (b) Camera tracking and local-to-global loop closure, ensuring robust pose estimation in different scenes; and (c) Inter-submap online distillation, which enforces geometric and photometric coherence in overlapping regions. This mechanism effectively eliminates boundary discontinuities, resulting in a seamless, high-fidelity global dense map consistent with the optimized global topology.}
\vspace{-0.4cm}
    \label{fig:system}
\end{figure*}

\section{Related work}
\subsection{Traditional SLAM}

Traditional visual SLAM systems have matured significantly, primarily splitting into sparse and dense methods. Sparse frameworks like ORB-SLAM3 \cite{orbslam3}, VINS-Mono \cite{vins} and so on~\cite{deng,xie} excel in real-time pose estimation and long-term localization via keyframe-based bundle adjustment. However, they lack the dense geometric context required for robotic interaction. Dense SLAM approaches, such as KinectFusion \cite{kinectfusion} and ElasticFusion \cite{elasticfusion}, provide complete surface reconstructions but often suffer from high memory overhead and sensitivity to tracking drift in large-scale environments. DTAM~\cite{dtam}, and CodeSLAM~\cite{codeslam} have further progressed to achieve dense representation of scenes by separating pose and depth estimation and introducing neural network-based estimation architectures. While systems like DROID-SLAM \cite{droidslam} have recently leveraged deep optical flow to enhance tracking robustness, the challenge of maintaining a globally consistent and memory-efficient dense map persists.

\subsection{Neural Scene Representation}
Neural network-based scene representation and reconstruction is another popular methodology in implicit environmental mapping. The advent of Neural Radiance Fields (NeRF) \cite{nerf} has redefined scene representation by encoding volumetric density and color within coordinate-based neural networks. To improve efficiency, subsequent works introduced explicit-implicit hybrid structures. Instant-NGP \cite{muller2022instant} utilized multi-resolution hash grids to accelerate training, while DVGO \cite{sun2022direct} and TensoRF \cite{chen2022tensorf} employed voxel grids and tensor decomposition for faster convergence.  Despite its numerous shortcomings, such as that it requires ground-truth poses for training input and that the single MLP takes hours to converge, NeRF has served as the basis for many later methods including BARF~\cite{barf}, NoPe-NeRF~\cite{bian2023nope}, LocalRF~\cite{meuleman2023progressively}, and ProSGNeRF~\cite{deng2026prosgnerf}. Specifically, the introduction of tri-planes offered a compelling balance between memory efficiency and reconstruction quality. These representations provide a continuous, high-fidelity alternative to traditional voxels or surfels, yet their application in incremental, real-time SLAM requires careful management of model plasticity and spatial scale.

\subsection{NeRF-based SLAM}
Integrating NeRF into SLAM systems has become a vibrant research frontier. iMAP \cite{imap} first demonstrated real-time neural SLAM using a single MLP, while NICE-SLAM \cite{niceslam} introduced hierarchical feature grids to enable more detailed reconstructions. Recent state-of-the-art methods like Co-SLAM \cite{coslam} and ESLAM \cite{eslam} further optimized the balance between tracking speed and map fidelity. NeSLAM~\cite{neslam} uses a depth completion and denoising method to improve scene representation. However, most existing neural SLAM frameworks are optimized for room-scale scenes. When facing large-scale trajectories, they frequently encounter catastrophic forgetting and cumulative drift due to the lack of robust global loop closure and scalable map management. 
GO-SLAM \cite{goslam} improves global pose optimization, while PLGSLAM \cite{plgslam} introduces progressive local representations and local-to-global BA for large-scale scenes.

Our work builds upon PLGSLAM's progressive local scene representation and local-to-global optimization paradigm, while extending them into a more comprehensive large-scale neural SLAM framework. Specifically, MSN-SLAM incorporates dense flow-based tracking and foundation-model-based loop closure into the local-to-global optimization, and further introduces inter-submap online distillation to explicitly enforce geometric and photometric consistency across independently optimized submaps. These extensions enable the system to handle longer trajectories and more challenging large-scale indoor and outdoor environments.

\section{Method}
\subsection{System Overview}
MSN-SLAM is a high-performance, parallelized framework consisting of four asynchronous threads: camera tracking, progressive mapping, loop closure optimization, and inter-submap online distillation. As illustrated in Fig. \ref{fig:system}, the system decouples the SLAM problem into four core asynchronous threads: camera tracking, progressive mapping, loop closure optimization, and inter-submap online distillation. The novelty of MSN-SLAM primarily stems from the unified system design that effectively integrates several complementary techniques into a scalable neural SLAM pipeline, rather than from any individual algorithmic module.

Formally, the system processes a continuous stream of sensory data $\mathcal{I} = \{I_t, D_t\}_{t=1}^N$, where $I_t$ and $D_t$ denote the RGB image and the corresponding depth map at timestamp $t$, respectively. The depth map $D_t$ is estimated online from the RGB image using a Vision Transformer (ViT)-based monocular depth estimation network. These predicted depth maps serve as geometric priors for camera tracking and neural scene optimization throughout the proposed RGB  SLAM pipeline.The objective is to concurrently estimate the 6-DoF camera trajectory $\mathcal{T} = \{\mathbf{T}_t \in SE(3)\}_{t=1}^N$ and a global implicit representation $\mathcal{S}$ composed of $M$ local neural submaps $\{S_i\}_{i=1}^M$. The system operation can be formulated as a mapping $\mathcal{M}$:
\begin{equation}
    [\mathcal{T}, \mathcal{S}] = \mathcal{M}(I_{1:N}, D_{1:N} \mid \Theta, \Phi)
\end{equation}
where $\Theta$ represents the learnable parameters of the hierarchical neural submaps and $\Phi$ denotes the state of the global keyframe database.

\subsection{Progressive Multi-submap Representation}

To achieve high-fidelity reconstruction while maintaining computational efficiency in large-scale robotic environments, we propose a progressive scene representation based on a collection of local neural submaps. Each submap $S_i$ encodes the local geometry and appearance using a hybrid architecture that combines explicit feature tri-planes and an implicit Multi-Layer Perceptron (MLP).

\noindent \textbf{Hybrid Neural Representation}
Each submap $S_i$ within the bounded volume $\Omega_i \subset \mathbb{R}^3$ is parameterized by a hybrid implicit-explicit representation. Specifically, we utilize axis-aligned tri-planes $\mathcal{V} = \{\mathbf{V}_{xy}, \mathbf{V}_{xz}, \mathbf{V}_{yz}\}$ to store high-frequency spatial features, complemented by a lightweight MLP $f_\theta$ to decode these features into geometric and photometric properties. Given a 3D query point $\mathbf{x} = (x, y, z)$ within the local coordinate system of submap $S_i$, we project $\mathbf{x}$ onto three orthogonal planes. The projected coordinates $\mathbf{x}_p \in \mathbb{R}^2$ are defined as:
\begin{equation}
    \mathbf{x}_{xy} = (x, y), \quad \mathbf{x}_{xz} = (x, z), \quad \mathbf{x}_{yz} = (y, z)
\end{equation}
The feature vector $\mathbf{f}_p$ for each projection is retrieved from the corresponding plane $\mathbf{V}_p$ via bilinear interpolation $\phi(\cdot, \cdot)$:
\begin{equation}
    \mathbf{f}_{xy} = \phi(\mathbf{V}_{xy}, \mathbf{x}_{xy}), \quad \mathbf{f}_{xz} = \phi(\mathbf{V}_{xz}, \mathbf{x}_{xz}), \quad \mathbf{f}_{yz} = \phi(\mathbf{V}_{yz}, \mathbf{x}_{yz})
\end{equation}
where $\mathbf{V}_p \in \mathbb{R}^{N \times N \times C}$ is a feature grid with resolution $N$ and channel dimension $C$. The local features from the three planes are aggregated to form a comprehensive descriptor $\mathbf{F}(\mathbf{x})$. To maximize the expressive power of the representation, we employ a concatenation strategy:
$\mathbf{F}(\mathbf{x}) = \mathbf{f}_{xy} \oplus \mathbf{f}_{xz} \oplus \mathbf{f}_{yz}$
where $\oplus$ denotes the concatenation operator. 

The aggregated feature $\mathbf{F}(\mathbf{x})$ is fed into the MLP $f_\theta$ to predict the volume density $\sigma \in [0, 1]$ and the RGB color $\mathbf{c} \in [0, 1]^3$:
\begin{equation}
    [\sigma(\mathbf{x}), \mathbf{c}(\mathbf{x})] = f_\theta(\mathbf{F}(\mathbf{x}), \gamma(\mathbf{x}))
\end{equation}
where $\gamma(\mathbf{x})$ is the positional encoding. 

Compared to traditional voxel-based representations where the memory footprint grows cubically ($O(N^3)$), the proposed tri-plane structure restricts the complexity to $O(N^2)$. This reduction is critical for large-scale SLAM, as it allows each submap to maintain high resolution without depleting the available GPU memory on embedded robotic platforms.

\noindent\textbf{Progressive Submap Initialization}
The system begins with an initial submap $S_1$. As the robot explores the environment, the system monitors the state of the active submap $S_{active}$. To prevent the neural model from over-fitting or losing representation density in new areas, we implement a progressive initialization strategy. A new submap $S_{i+1}$ is dynamically instantiated when the current camera pose $\mathbf{T}_t$ satisfies the following spatial and perceptual criteria:
\begin{equation}
    \mathcal{C}_{trigger} = \mathbb{I} \left( \|\mathbf{p}_t - \mathbf{c}_i\|_2 > \delta_{dist} \lor \frac{|\mathcal{K}_i \cap \mathcal{V}_t|}{|\mathcal{K}_i|} < \delta_{vis} \right)
\end{equation}
where $\mathbf{p}_t$ is the current camera position, $\mathbf{c}_i$ is the center of the active submap $S_i$, $\mathcal{K}_i$ is the set of keyframes associated with $S_i$, and $\mathcal{V}_t$ represents the current frustum's visible region. $\delta_{dist}$ and $\delta_{vis}$ are predefined thresholds for Euclidean distance and view overlap, respectively. Upon triggering, $S_{active}$ is set to the new submap $S_{i+1}$, while the parameters of $S_i$ are frozen and transferred to the global optimization thread. This progressive mechanism ensures that each local neural representation maintains its reconstruction fidelity without the interference of distant scene geometry.

\noindent \textbf{Differentiable Volume Rendering}
For each input image $I$, we utilize the camera intrinsics and the estimated camera pose to generate a set of rays $\mathbf{r}(t) = \mathbf{o} + t\mathbf{d}$. Along each ray, we sample $N$ points $\{\mathbf{x}_i\}_{i=1}^N$. We query our local neural submap $F_{\Theta}$ to retrieve the color $\mathbf{c}_i$ and volume density $\sigma_i$ at each sampled location: $F_{\Theta}:(\mathbf{x}_i, \mathbf{d}_i) \rightarrow (\mathbf{c}_i, \sigma_i)$. The final pixel color $\hat{\mathbf{C}}(\mathbf{r})$ is rendered by accumulating these values along the ray:
\begin{equation}
\begin{aligned}
    \hat{\mathbf{C}}(\mathbf{r}) & = \sum_{i=1}^N T_i \left(1 - \exp(-\sigma_i \delta_i)\right) \mathbf{c}_i, \\
    T_i & = \exp\left(-\sum_{j=1}^{i-1} \sigma_j \delta_j\right)
\end{aligned}
\end{equation}
where $\delta_i = t_{i+1} - t_i$ is the distance between adjacent sample points, and $T_i$ represents the accumulated transmittance along the ray. 

To enable the reconstruction of large-scale, unbounded environments, we employ a scene re-parameterization strategy. We apply a contraction function to map coordinates from Euclidean space into a bounded cubic volume:
\begin{equation}
    \operatorname{contract}(\mathbf{x}) = 
    \begin{cases}
        \mathbf{x} & \text{if } \|\mathbf{x}\|_{\infty} \leq 1 \\
        \left(2 - \frac{1}{\|\mathbf{x}\|_{\infty}}\right) \left(\frac{\mathbf{x}}{\|\mathbf{x}\|_{\infty}}\right) & \text{otherwise.}
    \end{cases}
\end{equation}
This warping mechanism allows the neural representation to capture distant background geometry with finite grid resolution, which is essential for outdoor robotic navigation.



\subsection{Local-to-Global Camera Tracking}

The tracking module ensures robust 6-DoF pose estimation and maintains a globally consistent keyframe-graph $(\mathcal{V}, \mathcal{E})$. Our system extends the tracking architecture of DROID-SLAM~\cite{droidslam} by introducing a \textit{local-to-global bundle adjustment} scheme driven by foundation-model-based loop closure.

\noindent\textbf{Dense Bundle Adjustment Layer}
For each incoming video stream, we apply a recurrent update operator based on RAFT \cite{raft} to compute the optical flow between the current frame and the last keyframe. If the average flow exceeds a threshold $\tau_{flow}$, a new keyframe is initialized. The poses $\mathbf{T} \in SE(3)$ and inverse depths $\mathbf{d} \in \mathbb{R}_{+}^{H \times W}$ are optimized using a differentiable Dense Bundle Adjustment (DBA) layer. Specifically, for each edge $(i, j) \in \mathcal{E}$ in the keyframe-graph, we minimize the following cost function:
\begin{equation}
    E(\mathbf{T}, \mathbf{d}) = \sum_{(i,j) \in \mathcal{E}} \| \mathbf{p}_{ij}^* - \Pi_{c}(\mathbf{G}_{ij} \circ \Pi_{c}^{-1}(\mathbf{p}_i, \mathbf{d}_i)) \|_{\Sigma_{ij}}^2
\end{equation}
where $\mathbf{p}_{ij}^*$ is the predicted flow from the recurrent operator, $\mathbf{w}_{ij}$ is the associated confidence map, and $\Sigma_{ij} = \text{diag}(\mathbf{w}_{ij})$. $\Pi_{c}$ and $\Pi_{c}^{-1}$ denote the projection and back-projection functions. The optimization utilizes a damped Gauss-Newton algorithm to find the optimal poses and depths of the local keyframes.

\noindent\textbf{Hierarchical Co-visibility and Foundation-Model-Based Loop Detection}
In our system, the construction of visual constraints in the keyframe-graph $(\mathcal{V}, \mathcal{E})$ is performed through a hierarchical co-visibility strategy. For the local sliding window, we utilize the mean rigid flow between keyframe pairs. However, while rigid-flow-based co-visibility is effective for local tracking, its computational complexity $O(N_{local} \times N_{KF})$ becomes prohibitive for global operations in large-scale scenarios. Furthermore, geometric flow estimation often becomes unstable in expansive outdoor environments due to sparse textures and significant viewpoint changes.

To address these limitations, we employ the SALAD descriptor~\cite{salad}, a foundation-model-based visual descriptor from DINO v2~\cite{dinov2}, for global loop closure detection. For each keyframe $KF_k$, we extract its invariant descriptor $\mathbf{g}_k = \Psi_{SALAD}(KF_k)$. Loop candidates are identified by querying the global database for historical keyframes that maximize the similarity.
By utilizing SALAD, we effectively reduce the number of redundant edges and optimization variables compared to the standard process.

\noindent \textbf{Local-to-Global Bundle Adjustment} Once a loop closure is detected, our system triggers a BA optimization process to ensure long-term consistency. Specifically, we employ a decoupled yet synchronized strategy: camera poses are refined through a differentiable Dense Bundle Adjustment (DBA) layer, while the neural scene representation is updated by minimizing a set of  objective functions.

For camera pose tracking thread, we insert long-range constraints into the keyframe-graph. The system performs a global DBA to minimize the alignment error between predicted and geometric optical flow. This propagates loop constraints throughout the trajectory. For mapping thread, we propose a composite loss function to ensure geometric accuracy and photometric fidelity:
\textit{Normalized Depth Loss:} To supervise the geometry, we render the depth map $\hat{\mathbf{D}}$ using the current scene representation:
\begin{equation}
\hat{\mathbf{D}}(\mathbf{r})=\sum_{i=1}^N T_i\left(1-\exp \left(-\sigma_i \delta_i\right)\right) d_i
\end{equation}
Since monocular depth estimates are not scale- and shift-invariant, we normalize the rendered depth $\hat{\mathbf{D}}$ to align it with the prior from our depth network. We estimate the scale $s(D)$ and shift $t(D)$ as follows:
$t(D)=\frac{1}{N}\sum_{i=1}^{N}D, \quad s(D)=\frac{1}{N}\sum_{i=1}^{N}|D-t(D)|$
The normalized depth loss $\mathcal{L}_d$ is defined by the difference between the normalized rendered depth $\hat{\mathbf{D}}^*$ and the normalized estimated depth $\mathbf{D}^*$:
\begin{equation}
\mathcal{L}_d=\left|\hat{\mathbf{D}}^*-\mathbf{D}^*\right|, \quad \text{where } \mathbf{D}^* = \frac{D - t(D)}{s(D)}
\end{equation}

\textit{Photo-metric Loss:} To optimize the appearance and local radiance fields, we minimize the color difference:
$
\mathcal{L}_p=\|\hat{\mathbf{C}}(\mathbf{r})-\mathbf{C}(\mathbf{r})\|_2^2
$
where $\hat{\mathbf{C}}(\mathbf{r})$ is the rendered color and $\mathbf{C}(\mathbf{r})$ is the input RGB image.

\textit{Optical Flow Loss:} To enhance robustness in dynamic or challenging scenarios, we introduce induced optical flow constraints. The forward flow $\hat{\mathcal{F}}_{k \rightarrow k+1}$ is computed via back-projection:
\begin{equation}
\hat{\mathcal{F}}_{k \rightarrow k+1}=(u, v)-\Pi\left(\{R,\mathbf{t}\}_{k \rightarrow k+1} \Pi^{-1}(u, v, \hat{D})\right)
\end{equation}
The forward and backward flow losses ($\mathcal{L}_{fa}, \mathcal{L}_{fb}$) ensure that the implicit scene representation remains consistent with the observed motion dynamics:
\begin{equation}
\mathcal{L}_{fa}=\left\|\hat{\mathcal{F}}_{k \rightarrow k+1}-\mathcal{F}_{k \rightarrow k+1}\right\|_1
\end{equation}

\subsection{Inter-Submap Online Distillation}
While global Bundle Adjustment (BA) ensures the topological consistency of the camera trajectory, the independent optimization of local neural submaps may still lead to geometric misalignments or "seam" artifacts in overlapping regions. To achieve a seamless global dense reconstruction, we propose an \textit{Inter-Submap Online Distillation} algorithm that enforces spatial and photometric coherence between adjacent submaps.

\noindent \textbf{Overlap Region Sampling}
For any two adjacent submaps $S_i$ and $S_j$ with learnable parameters $\Theta_i$ and $\Theta_j$, we first identify their spatial overlap region $\Omega_{ij} = \Omega_i \cap \Omega_j$. To maintain real-time performance, we do not perform dense volumetric sampling over the entire intersection. Instead, we focus on the boundaries where the camera transitions between submaps. Specifically, we sample a set of 3D points $\mathcal{X}_{overlap} = \{\mathbf{x}_k \in \Omega_{ij}\}$ using a probability density function guided by the visibility of the keyframes associated with both submaps.

\noindent \textbf{Teacher-Student Consistency Constraint}
During the online distillation process, the submap with more historical observations or higher optimization maturity (typically the earlier submap $S_i$) acts as the \textit{Teacher}, providing geometric and photometric supervision to the newly initialized \textit{Student} submap $S_j$. 
For each sampled point $\mathbf{x} \in \mathcal{X}_{overlap}$, we query the occupancy probability $o$ and the RGB color $\mathbf{c}$ from both submaps. We define the distillation loss $\mathcal{L}_{distill}$ to minimize the divergence between the two neural representations:

\begin{equation}
    \mathcal{L}_{distill} = \sum_{\mathbf{x} \in \mathcal{X}_{overlap}} \left( \omega_g \|o_i(\mathbf{x}) - o_j(\mathbf{x})\|^2 + \omega_c \|\mathbf{c}_i(\mathbf{x}) - \mathbf{c}_j(\mathbf{x})\|^2 \right)
\end{equation}
where $\omega_g$ and $\omega_c$ are weighting coefficients for geometric and photometric consistency, respectively. By enforcing this constraint, the system ensures that the "Student" submap $S_j$ inherits the established structural information from $S_i$, effectively eliminating ghosting artifacts and surface discontinuities.


\section{Experimental Results}
\subsection{Implementation Details}
In the tracking thread, the local sliding window size is set to $N_{local}=15$ keyframes for image sequences. The optical flow threshold for keyframe creation is $\tau_{flow}=2.5$ pixels. For the SALAD-based loop closure, we extract $544$ dimensional descriptors for each keyframe and set the similarity threshold to $\tau_{co}=0.75$. The submap initialization is triggered when the camera displacement exceeds $\delta_{dist}=2.0$m or the visibility ratio drops below $\delta_{vis}=0.6$.

\noindent \textbf{Baselines Methods and Public Datasets.} The main baseline methods we compared are BARF~\cite{barf}, NoPe-NeRF~\cite{nopenerf},  LocalRF~\cite{progressive} and Flow-NeRF~\cite{flownerf}. All of them are joint learning methods of camera poses and implicit scene representation. We also compare our approach with several state-of-the-art SLAM methods, including NICE-SLAM~\cite{niceslam} and GO-SLAM~\cite{goslam}. We evaluate the scalability and robustness of our framework across a diverse range of environments, spanning from room-scale interiors to city-scale urban trajectories:
\begin{itemize}
    \item Small-to-Medium Scale Environments: We utilize the Tanks and Temples dataset (ranging from $4\times4$m to $8\times8$m), and the Static Hikes dataset \cite{progressive} ($15\times15$m) to assess reconstruction fidelity and the stability of the neural submap representation in both indoor and outdoor settings.
    \item City-Scale Urban Scenes: To verify the effectiveness of our local-to-global loop closure and submap management in expansive areas, we conduct tests on the KITTI \cite{kitti}  benchmark. These sequences cover approximately $500\text{m} \times 400\text{m}$, representing a rigorous test for long-term consistency and drift suppression in real-world urban reconstruction.
\end{itemize}
The baseline methods are selected primarily according to their applicability to large-scale or long-sequence scene reconstruction and the similarity of their input and evaluation settings to ours. For example, NoPe-NeRF and LocalRF provide closely related references for scalable scene representation and image-based reconstruction. By contrast, ESLAM is mainly evaluated on room-scale indoor RGB-D benchmarks and directly relies on measured depth for TSDF supervision and tracking, resulting in a less closely matched experimental setting.

\subsection{System Integration and Sensing Platform}
To bridge the gap between benchmark evaluation and practical robotic deployment, we developed a specialized handheld mechatronic sensory platform for comprehensive system validation. While public datasets provide a standardized basis for comparison, our physical platform allows for the assessment of the proposed algorithm  under real-world scenarios.

\begin{figure}[ht]
    \centering
    
    \includegraphics[width=\linewidth]{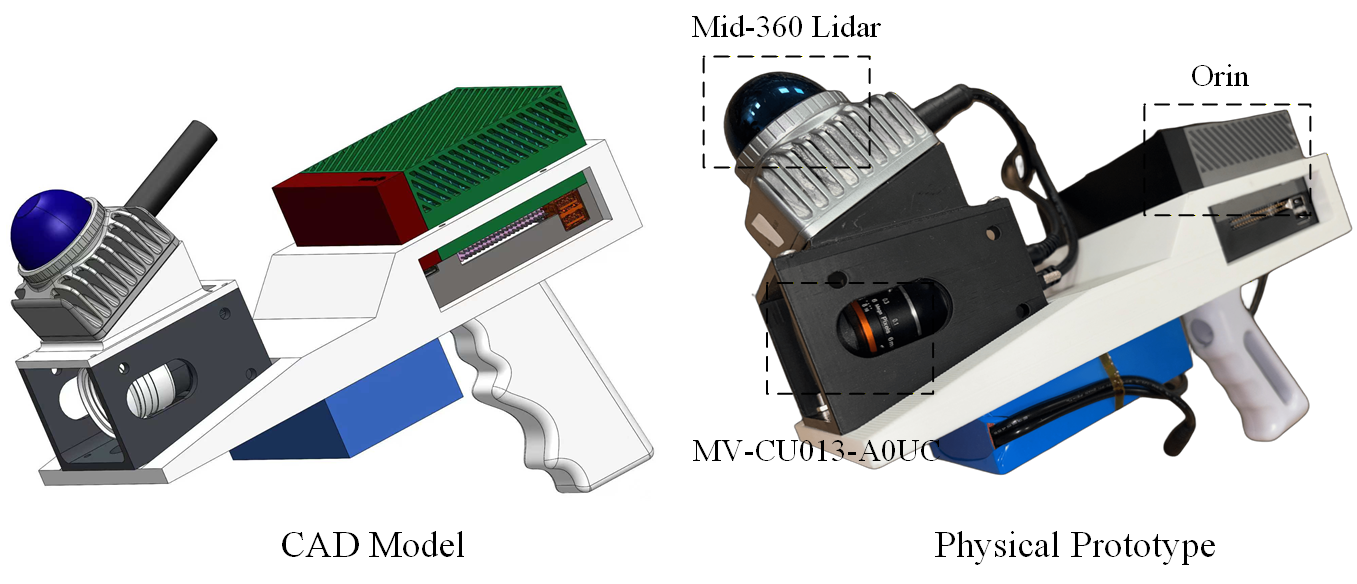}
\vspace{-0.4cm}
    \caption{Overview of the proposed handheld device: (a) CAD model showing the internal structural layout; (b) Physical prototype of the assembled device.}
\vspace{-0.4cm}
    \label{fig:cad}
\end{figure}

As illustrated in Fig. \ref{fig:cad}, the platform is engineered as a high-performance integrated system. The sensing suite comprises:
\begin{itemize}
    \item Livox Mid-360 LiDAR: A high-performance LiDAR sensor that provides a $360^\circ \times 59^\circ$ field-of-view (FOV), enabling the acquisition of dense and precise 3D geometric structures in expansive environments. The LiDAR is primarily employed to provide synchronized geometric references for the construction and evaluation of the real-world dataset, while the online reconstruction pipeline relies solely on RGB inputs. This configuration enables fair comparison with existing neural RGB  SLAM methods while maintaining a practical multi-sensor platform for robotic deployment.
    \item Industrial-grade Global Shutter Camera MV-CU013-A0UC: To ensure high-fidelity visual perception, an industrial camera is utilized to capture synchronized image streams. 
    \item NVIDIA Jetson AGX Orin: The central computing unit is a high-performance embedded module capable of 275 TOPS of AI performance. All core tracking and submap management modules are deployed on this onboard unit to verify the real-time operational feasibility of our neural SLAM framework in power-constrained scenarios.
\end{itemize}

To ensure the structural integrity and precise spatial-temporal synchronization of the multi-sensor system, we designed a customized integrated housing module fabricated via high-strength 3D printing technology. Drawing inspiration from the robust synchronization architecture in FAST-LIO2 \cite{fastlivo2}, we implemented a hardware-level time synchronization protocol. By utilizing a common clock source and hardware triggers, the timestamps of the industrial camera's exposure and the Mid-360's point cloud packets are strictly aligned with sub-millisecond precision. This temporal consistency, combined with the rigid 3D-printed mounting, maintains stable extrinsic calibration even during high-frequency jitter, significantly reducing the measurement innovations' residual in our joint optimization framework.

The entire platform was subjected to testing across a variety of challenging real-world scenarios, including high-ceiling lobbies, low-texture long corridors, and transitional indoor-outdoor zones, validating the system's robustness and scalability.

\begin{figure*}[ht]
    \centering
    
    \includegraphics[width=0.9\linewidth]{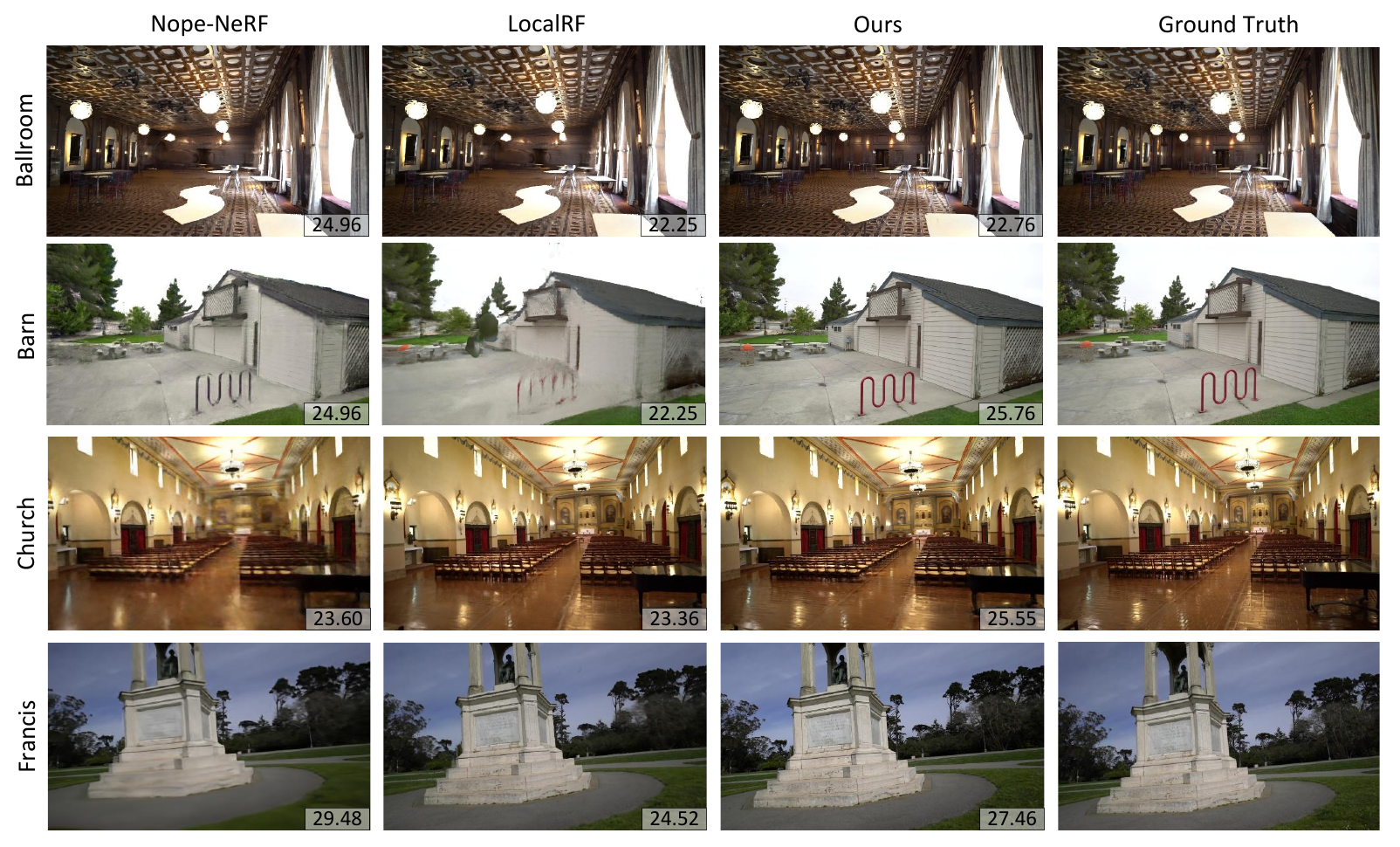}
\vspace{-0.4cm}
    \caption{Visual quality comparison on Tanks and Temples sequences. Compared to SOTA methods including NoPe-NeRF~\cite{nopenerf} and LocalRF~\cite{progressive}, our method provides more complete and visually consistent 3D models. The PSNR metric for each rendering is shown in the bottom-right corner, highlighting the quantitative and qualitative improvements of our framework in diverse environments.}
\vspace{-0.4cm}
    \label{fig:tanks}
\end{figure*}

\begin{table*}[]
\caption{Novel view synthesis results on Tanks and Temples~\cite{tanks}. We use PSNR, SSIM, and LPIPS as our metrics.}
\vspace{-0.2cm}
\resizebox{\textwidth}{!}{
\begin{tabular}{ccccccccccccccccc}
\hline
\multicolumn{2}{c}{\multirow{2}{*}{Scenes}} & \multicolumn{3}{c}{BARF} & \multicolumn{3}{c}{NoPe-NeRF} & \multicolumn{3}{c}{LocalRF} & \multicolumn{3}{c}{Flow-NeRF} & \multicolumn{3}{c}{Ours} \\ \cline{3-17}
\multicolumn{2}{c}{} 
& PSNR$\uparrow$ & SSIM$\uparrow$ & LPIPS$\downarrow$
& PSNR$\uparrow$ & SSIM$\uparrow$ & LPIPS$\downarrow$
& PSNR$\uparrow$ & SSIM$\uparrow$ & LPIPS$\downarrow$
& PSNR$\uparrow$ & SSIM$\uparrow$ & LPIPS$\downarrow$
& PSNR$\uparrow$ & SSIM$\uparrow$ & LPIPS$\downarrow$ \\ \hline

\multirow{8}{*}{Tanks \& Temples} 
& Ballroom & 20.66 & 0.50 & 0.60 & 24.96 & 0.71 & 0.39 & 22.25 & 0.75 & 0.26 & \textbf{28.83} & \textbf{0.86} & 0.24 & 26.50 & 0.85 & \textbf{0.08} \\
& Barn     & 25.28 & 0.64 & 0.48 & 25.57 & 0.68 & 0.44 & 21.28 & 0.68 & 0.36 & 28.53 & 0.78 & 0.35 & \textbf{28.59} & \textbf{0.91} & \textbf{0.11} \\
& Church   & 23.17 & 0.62 & 0.52 & 23.6  & 0.67 & 0.46 & 23.36 & 0.77 & 0.19 & 28.27 & 0.83 & 0.28 & \textbf{28.82} & \textbf{0.92} & \textbf{0.08} \\
& Family   & 23.04 & 0.61 & 0.56 & 23.77 & 0.68 & 0.48 & 24.24 & 0.80 & 0.17 & 29.40 & 0.85 & 0.29 & \textbf{30.48} & \textbf{0.94} & \textbf{0.05} \\
& Francis  & 25.82 & 0.69 & 0.57 & 29.48 & 0.80 & 0.38 & 24.52 & 0.74 & 0.28 & 30.63 & 0.83 & 0.33 & \textbf{33.38} & \textbf{0.93} & \textbf{0.10} \\
& Horse    & 24.09 & 0.72 & 0.41 & 25.00 & 0.83 & 0.27 & 23.44 & 0.79 & 0.23 & 28.57 & 0.86 & 0.23 & \textbf{30.07} & \textbf{0.93} & \textbf{0.07} \\
& Ignatius & 21.78 & 0.47 & 0.60 & 23.77 & 0.61 & 0.47 & 21.66 & 0.63 & 0.33 & 26.25 & 0.73 & 0.35 & \textbf{27.83} & \textbf{0.87} & \textbf{0.09} \\
& Museum   & 23.58 & 0.61 & 0.55 & 25.26 & 0.76 & 0.36 & 21.93 & 0.67 & 0.35 & \textbf{29.43} & 0.85 & 0.27 & \textbf{29.43} & \textbf{0.92} & 0.12 \\

\multicolumn{2}{c}{\textbf{Mean}} 
& 23.43 & 0.61 & 0.54 
& 25.18 & 0.72 & 0.41 
& 22.84 & 0.73 & 0.27 
& 28.73 & 0.82 & 0.29 
& \textbf{29.84} & \textbf{0.90} & \textbf{0.09} \\ \hline
\end{tabular}
\vspace{-0.8cm}
}

\label{tab:tanks}
\end{table*}

\begin{figure*}[ht]
    \centering
    
    \includegraphics[width=0.9\linewidth]{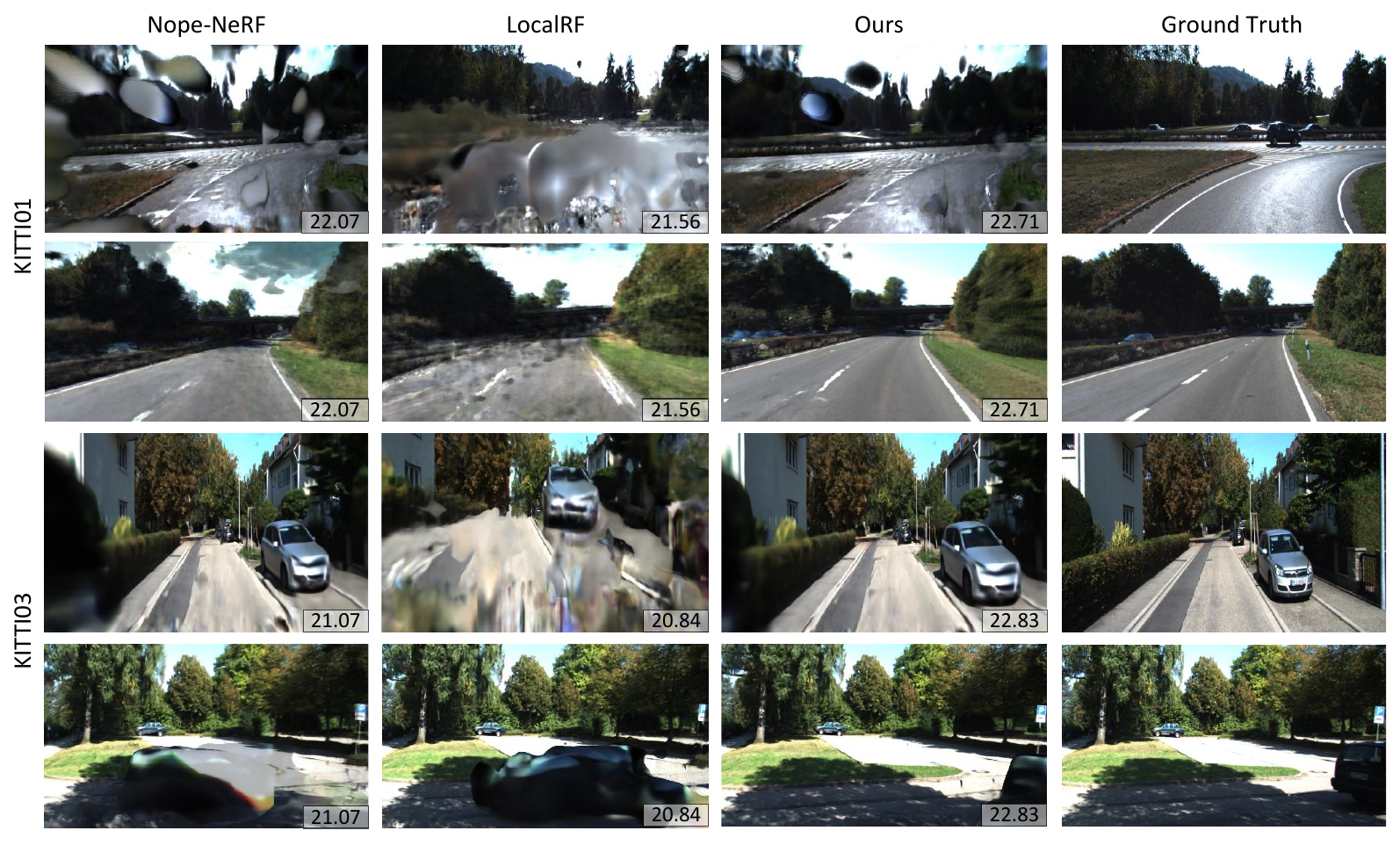}
\vspace{-0.4cm}
    \caption{Visual quality comparison on the city-scale KITTI dataset. Compared to SOTA methods including NoPe-NeRF and LocalRF, our method provides more complete and visually consistent 3D models in expansive urban environments (approx. $500\text{m} \times 400\text{m}$). The PSNR metrics highlight the robust structural integrity maintained by our Progressive Multi-submap strategy, local-to-global loop closure and BA over long-duration trajectories.}
\vspace{-0.4cm}
    \label{fig:kitti}
\end{figure*}
\subsection{Evaluation Protocol}
For the real-world experiments, the reference trajectory is constructed from the RGB input sequence using COLMAP. The estimated trajectories are aligned with the reference trajectory before computing Absolute Trajectory Error (ATE) and Relative Pose Error (RPE). Both metrics follow the standard evaluation protocol widely adopted in visual SLAM literature. Reconstruction quality is evaluated using PSNR, SSIM, and LPIPS under the same protocol as the compared neural SLAM methods.

\subsection{Scene Reconstruction}
We evaluate the geometric and photometric fidelity of our multi-submap representation. Our system's ability to maintain high-resolution details while scaling to expansive environments is benchmarked against several neural SLAM baselines.

\noindent \textbf{High-Fidelity Reconstruction}
The Tanks and Temples dataset  provides a rigorous test for high-fidelity reconstruction in both indoor and outdoor settings. 
As reported in Table~\ref{tab:tanks} and Fig.~\ref{fig:tanks}, our method achieves strong overall view-synthesis performance while preserving fine-grained structural details across the evaluated scenes. This improvement benefits from the hybrid neural representation and inter-submap online distillation.

\noindent \textbf{Scalability to Large-Scale and Long-sequence Scenes}
To validate the effectiveness of our Progressive Multi-Submap strategy and Local-to-Global BA, we conduct extensive tests on the KITTI~\cite{kitti} and Static Hikes~\cite{progressive} datasets, which represent city-scale ($500\text{m} \times 400\text{m}$) and medium-scale environments, respectively.

On the KITTI dataset demonstrated in Fig.~\ref{fig:kitti}, the combination of global loop closure and progressive multi-submap mapping maintains competitive trajectory accuracy over long sequences, as summarized in Table~\ref{tab:pose2}.

\begin{figure*}[ht]
    \centering
    
    \includegraphics[width=0.9\linewidth]{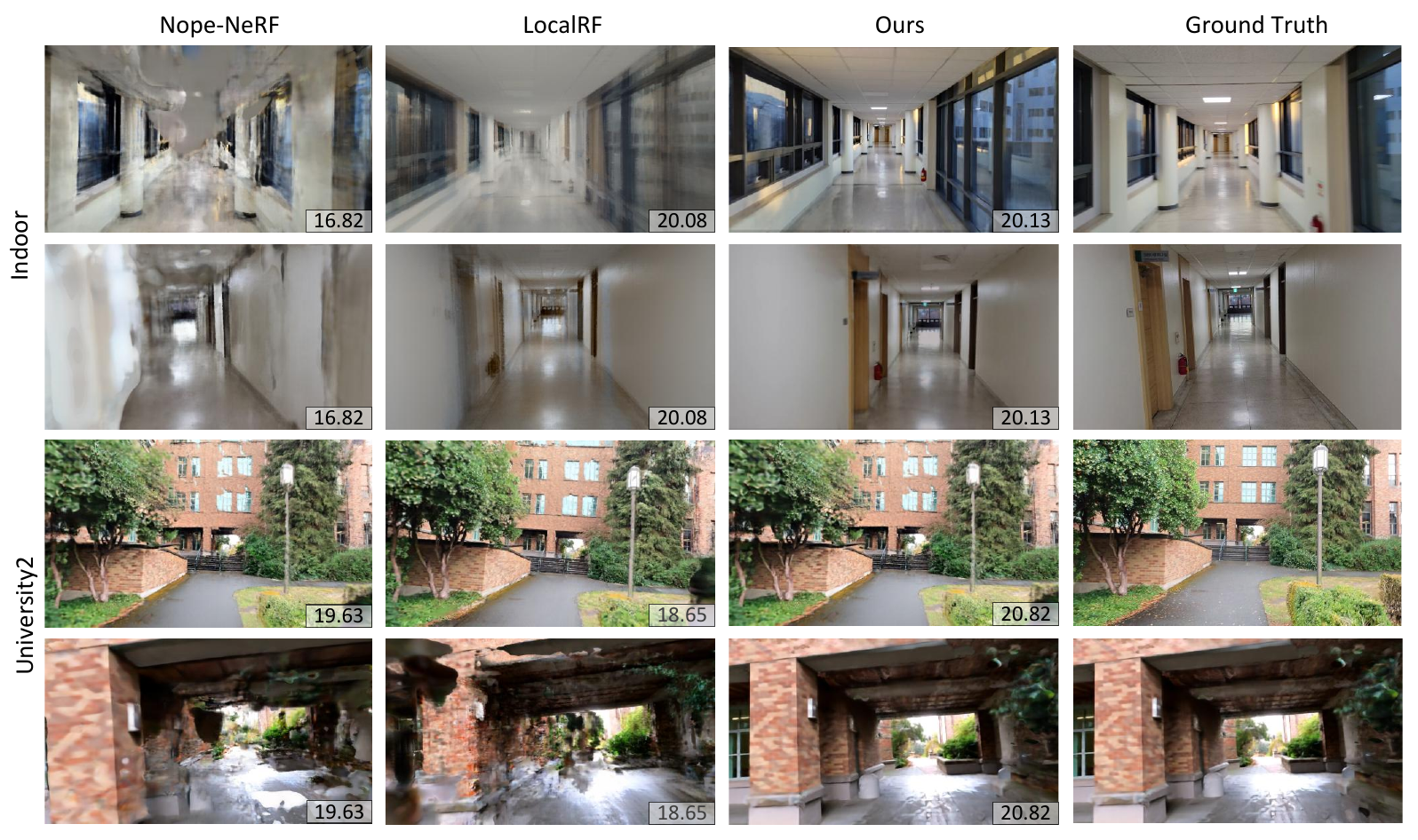}
\vspace{-0.4cm}
    \caption{Reconstruction results on the Static Hikes dataset~\cite{progressive} . Our Progressive Multi-Submap strategy enables the efficient management of memory resources while capturing distant background geometry through the contract function. The joint optimization of poses and neural parameters and online distillation facilitates a superior balance between tracking robustness and dense mapping fidelity.}

    \label{fig:hikes}
    \vspace{-0.4cm}
\end{figure*}

The results on Static Hikes in Fig.~\ref{fig:hikes} demonstrate the robustness of our Progressive Multi-Submap strategy. By re-parameterizing unbounded spaces into bounded neural volumes via the Contract Function, our method successfully reconstructs distant background geometry that other SOTA methods typically fail to represent.

\begin{table*}[]
\caption{Novel view synthesis results on Static Hikes~\cite{progressive} and KITTI~\cite{kitti}. We use PSNR, SSIM, and LPIPS as our metrics.}
\vspace{-0.3cm}
\resizebox{\textwidth}{!}{
\begin{tabular}{cccccccccccccc}
\hline
\multicolumn{2}{c}{\multirow{2}{*}{Scenes}} & \multicolumn{3}{c}{BARF} & \multicolumn{3}{c}{NoPe-NeRF} & \multicolumn{3}{c}{LocalRF} & \multicolumn{3}{c}{Ours}                       \\ \cline{3-14} 
\multicolumn{2}{c}{}                        & PSNR$\uparrow$                 & SSIM$\uparrow$                  & LPIPS$\downarrow$ & PSNR$\uparrow$                 & SSIM$\uparrow$                  & LPIPS$\downarrow$    & PSNR$\uparrow$                 & SSIM$\uparrow$                  & LPIPS$\downarrow$   & PSNR$\uparrow$                 & SSIM$\uparrow$                  & LPIPS$\downarrow$         \\ \hline
\multirow{2}{*}{Static Hike}  & Indoor      & 14.81   & 0.69  & 0.74   & 16.82    & 0.74    & 0.62     & 20.08    & 0.70   & 0.45    & \textbf{20.13} & \textbf{0.80} & \textbf{0.45} \\
                              & University2 & 11.39   & 0.29  & 0.88   & 19.63    & 0.49    & 0.63     & 18.65    & 0.43   & 0.42    & \textbf{20.82} & \textbf{0.67} & \textbf{0.37} \\
Ward                          & Ward        & 10.04   & 0.33  & 0.79   & 24.28    & 0.81    & 0.45     & 23.11    & 0.82   & 0.29    & \textbf{24.86} & \textbf{0.84} & \textbf{0.24} \\
\multirow{2}{*}{KITTI}        & KITTI01     & 11.49   & 0.67  & 0.42   & 22.07    & 0.77    & 0.31     & 21.56    & 0.76   & 0.33    & \textbf{22.71} & \textbf{0.78} & \textbf{0.30} \\
                              & KITTI03     & 12.28   & 0.56  & 0.75   & 22.54    & 0.74    & 0.33     & 20.78    & 0.75   & 0.34    & \textbf{22.83} & \textbf{0.78} & \textbf{0.29} \\
\multicolumn{2}{c}{\textbf{Mean}}           & 12.00   & 0.51  & 0.72   & 21.07    & 0.71    & 0.47     & 20.84    & 0.69   & 0.37    & \textbf{22.27} & \textbf{0.77} & \textbf{0.33} \\ \hline
\end{tabular}
}
\label{tab:kitti}
\vspace{-0.4cm}
\end{table*}

\subsection{Pose Estimation}
The localization performance of our system is evaluated using the Absolute Trajectory Error (ATE) RMSE [cm]. The results demonstrate that our hierarchical optimization strategy effectively suppresses cumulative drift, ensuring high-precision navigation even in expansive environments.

We report rotational and translational Relative Pose Error as $RPE_{\textit{r}}$ and $RPE_{\textit{t}}$, respectively, following the standard trajectory evaluation protocol.

As reported in Table~\ref{tab:pose1}, our method achieves the lowest average ATE among all compared methods, demonstrating strong overall localization accuracy. While these methods rely on local geometric consistency, our framework allows the neural representation to provide stable geometric priors, resulting in a more robust trajectory. This is attributed to our Local BA module, which efficiently refines poses within a sliding window, preventing the immediate divergence often seen in purely frame-to-model tracking frameworks.

In the KITTI sequences, where cumulative drift is a major bottleneck, our SALAD-driven loop closure successfully identifies long-range constraints that are often overlooked by the rigid-flow-based methods in DROID-SLAM and GO-SLAM. Because SALAD leverages foundation-model-based semantic features, it remains robust under the drastic viewpoint variations common in urban navigation. While monolithic neural SLAM methods often suffer from memory exhaustion in city-scale scenes, our Progressive Multi-Submap architecture, combined with Local-to-Global BA, ensures that the global keyframe-graph is refined only when necessary. This allows for the correction of large-scale drift without compromising the real-time performance of the local tracking thread. The Inter-Submap Online Distillation also contributes to pose accuracy by ensuring that the map remains geometrically consistent across submap transitions. This prevents the tracking thread from experiencing "jumps" or discontinuities when moving between different neural volumes, contributing to the lowest average ATE reported in Table~\ref{tab:pose2}.

\begin{table*}[]
\caption{Quantitative comparison of Absolute Trajectory Error (ATE) RMSE [cm] on the Tanks and Temples dataset~\cite{tanks}. We use ATE and RPE as our metrics.}
\resizebox{\textwidth}{!}{
\begin{tabular}{cccccccccccccc}
\hline
\multicolumn{2}{c}{\multirow{2}{*}{Scenes}}  & \multicolumn{3}{c}{BARF} & \multicolumn{3}{c}{NoPe-NeRF}                    & \multicolumn{3}{c}{DROID-SLAM} & \multicolumn{3}{c}{Ours}                         \\ \cline{3-14} 
\multicolumn{2}{c}{}                         & $ATE\downarrow$    & $RPE_{\textit{r}}\downarrow$  & $RPE_{\textit{t}}\downarrow$  & $ATE\downarrow$    & $RPE_{\textit{r}}\downarrow$  & $RPE_{\textit{t}}\downarrow$         & $ATE\downarrow$    & $RPE_{\textit{r}}\downarrow$  & $RPE_{\textit{t}}\downarrow$    & $ATE\downarrow$    & $RPE_{\textit{r}}\downarrow$  & $RPE_{\textit{t}}\downarrow$          \\ \hline
\multirow{8}{*}{Tanks \& Temples} & Ballroom & 0.019  & 0.228  & 0.343  & \textbf{0.003} & \textbf{0.026} & \textbf{0.054} & 0.142    & 1.419    & 1.441    & 0.010          & 0.152          & 0.231          \\
                                  & Barn     & 0.075  & 0.326  & 1.402  & 0.023          & \textbf{0.034} & \textbf{0.127} & 0.003    & 0.674    & 0.382    & \textbf{0.001} & 0.282          & 0.262          \\
                                  & Church   & 0.059  & 0.063  & 0.458  & 0.026          & \textbf{0.013} & \textbf{0.088} & 0.002    & 0.219    & 0.211    & \textbf{0.002} & 0.142          & 0.182          \\
                                  & Family   & 0.116  & 0.595  & 0.577  & 0.006          & \textbf{0.045} & \textbf{0.049} & 0.167    & 2.182    & 1.780    & \textbf{0.006} & 0.149          & 0.167          \\
                                  & Francis  & 0.095  & 0.749  & 0.924  & \textbf{0.005} & \textbf{0.029} & \textbf{0.069} & 0.001    & 0.644    & 0.910    & \textbf{0.001} & 0.044          & 0.116          \\
                                  & Horse    & 0.016  & 0.399  & 0.239  & 0.005          & \textbf{0.032} & 0.192          & 0.203    & 2.667    & 2.729    & 0.029          & 0.102          & \textbf{0.134} \\
                                  & Ignatius & 0.057  & 0.288  & 1.187  & 0.003          & \textbf{0.010} & \textbf{0.039} & 0.198    & 3.197    & 2.720    & \textbf{0.002} & 0.065          & 0.126          \\
                                  & Museum   & 0.257  & 1.128  & 2.589  & 0.035          & 0.247          & \textbf{0.335} & 0.150    & 1.213    & 1.753    & \textbf{0.024} & \textbf{0.162} & 0.445          \\
\multicolumn{2}{c}{\textbf{Mean}}            & 0.087  & 0.472  & 0.965  & 0.013          & \textbf{0.055} & \textbf{0.119} & 0.108    & 1.527    & 1.491    & \textbf{0.009} & 0.137          & 0.208          \\ \hline
\end{tabular}
}
\label{tab:pose1}
\end{table*}

\begin{table*}[]
\caption{Pose estimation accuracy (ATE RMSE [m]) on city-scale KITTI and medium-scale Static Hikes datasets. }
\resizebox{\textwidth}{!}{
\begin{tabular}{cccccccccccccc}
\hline
\multicolumn{2}{c}{\multirow{2}{*}{Scenes}} & \multicolumn{3}{c}{BARF} & \multicolumn{3}{c}{NoPe-NeRF}                    & \multicolumn{3}{c}{DROID-SLAM} & \multicolumn{3}{c}{Ours}                         \\ \cline{3-14} 
\multicolumn{2}{c}{}                        & $ATE\downarrow$    & $RPE_{\textit{r}}\downarrow$  & $RPE_{\textit{t}}\downarrow$  & $ATE\downarrow$    & $RPE_{\textit{r}}\downarrow$  & $RPE_{\textit{t}}\downarrow$          & $ATE\downarrow$    & $RPE_{\textit{r}}\downarrow$  & $RPE_{\textit{t}}\downarrow$          & $ATE\downarrow$    & $RPE_{\textit{r}}\downarrow$  & $RPE_{\textit{t}}\downarrow$          \\ \hline
\multirow{2}{*}{Static Hike}  & Indoor      & 0.632  & 0.545  & 1.791  & 0.463          & \textbf{0.443} & 1.595          & 0.203 & 2.053 & 0.752          & \textbf{0.068} & 0.635          & \textbf{0.645} \\
                              & University2 & 0.075  & 1.926  & 2.675  & 0.588          & \textbf{0.340} & 1.476          & 0.035 & 2.571 & 0.849          & \textbf{0.006} & 1.355          & \textbf{0.717} \\
Ward                          & Ward        & 1.142  & 0.278  & 3.449  & 0.839          & 0.611          & 4.997          & 0.786 & 1.550 & 4.101          & \textbf{0.677} & \textbf{0.214} & \textbf{3.433} \\
\multirow{2}{*}{KITTI}        & KITTI01     & 1.058  & 0.600  & 14.601 & 0.457          & \textbf{0.078} & 9.501          & 0.201 & 0.447 & 5.590          & \textbf{0.174} & 0.152          & \textbf{4.536} \\
                              & KITTI03     & 2.254  & 1.066  & 19.46  & \textbf{0.194} & \textbf{0.038} & \textbf{3.251} & 0.559 & 0.219 & 1.683          & 0.280          & 0.159          & 3.933          \\
\multicolumn{2}{c}{\textbf{Mean}}                    & 1.032  & 0.883  & 8.395  & 0.508          & \textbf{0.302} & 4.164          & 0.357 & 1.368 & \textbf{2.595} & \textbf{0.241} & 0.503          & 2.653          \\ \hline
\end{tabular}
}
\label{tab:pose2}
\vspace{-0.4cm}
\end{table*}

\subsection{Real-World Integrated Validation}
To further evaluate the robustness and real-time capability of our system in practical robotic applications, we conducted a series of experiments using our customized handheld mechatronic platform. Unlike controlled public datasets, these real-world sequences involve aggressive motion jitter, diverse lighting conditions, and complex structural layouts.

 We recorded four representative sequences across various campus environments, spanning from constrained indoor rooms to expansive outdoor-indoor transitional zones. The detailed statistics of these sequences are summarized in Table~\ref{tab:dataset}.
 
The real-world sequences cover confined rooms of different sizes, low-texture corridors, large open spaces, and indoor–outdoor transitions, as demonstrated in Table~\ref{tab:dataset}. The results demonstrate that the proposed tracking, progressive submap management, and inter-submap consistency mechanisms remain effective under these diverse operating conditions.

During all real-world tests, the tracking and mapping threads were executed entirely on the NVIDIA Jetson AGX Orin. The system achieved a consistent processing rate of 10 FPS. Temporally aligned sensor acquisition is ensured by hardware-level synchronization.

\begin{table}[t]

\caption{Statistics and Environmental Characteristics of Real-World Validation Sequences recorded by the Handheld Platform.}
\centering
\scalebox{0.75}{
\setlength{\tabcolsep}{0.3mm}{
\begin{tabular}{llcc p{5cm}} 
\toprule
\textbf{Sequence ID} & \textbf{Scenario Type} & \textbf{Length [m]} & \textbf{Area [m$^2$]} & \textbf{Main Challenges} \\
\midrule
Lab-Room   & Confined Office  & 25.4  & $6 \times 8$   & Cluttered objects, near-range geometry \\
Corridor   & Narrow Passage   & 85.0  & $2 \times 60$  & Low-texture walls, repetitive structures \\
Grand-Hall & Large Open Space & 120.5 & $25 \times 35$ & High-ceiling, sparse geometric features \\
Trans-IO   & Indoor-Outdoor   & 165.2 & Unbounded      & Drastic light changes, scale variation \\
\bottomrule
\end{tabular}}}
\label{tab:dataset}
\vspace{-0.4cm}
\end{table}

\begin{table}[t]
\caption{Comparative Results of Localization and Reconstruction on our Handheld Mechatronic Platform. The best results are highlighted in \textbf{bold}.}
\label{tab:real_world_comparison}
\centering
\resizebox{\columnwidth}{!}{ 
\begin{tabular}{ll ccc}
\toprule
Sequence & Method & ATE RMSE [cm] $\downarrow$ & PSNR [dB] $\uparrow$ & F-score [\%] $\uparrow$ \\
\midrule
\multirow{3}{*}{Lab-Room}   & NICE-SLAM \cite{niceslam} & 4.52  & 22.15 & 82.4 \\
                            & GO-SLAM \cite{goslam}    & 2.42  & 24.89 & 89.9 \\
                            & Ours       & \textbf{2.05} & \textbf{26.45} & \textbf{93.2} \\
\midrule
\multirow{3}{*}{Corridor}   & NICE-SLAM \cite{niceslam} & 15.64 & 19.84 & 75.2 \\
                            & GO-SLAM \cite{goslam}    & 4.95  & 21.56 & 81.6 \\
                            & Ours       & \textbf{4.12} & \textbf{23.82} & \textbf{89.4} \\
\midrule
\multirow{3}{*}{Grand-Hall} & NICE-SLAM \cite{niceslam} & 32.18 & 18.52 & 68.7 \\
                            & GO-SLAM \cite{goslam}    & 10.35 & 20.94 & 76.5 \\
                            & Ours       & \textbf{9.84} & \textbf{24.15} & \textbf{88.1} \\
\midrule
\multirow{3}{*}{Trans-IO}   & NICE-SLAM \cite{niceslam} & — (Fail) & 17.12 & 62.1 \\
                            & GO-SLAM \cite{goslam}    & 22.45 & 19.65 & 72.8 \\
                            & Ours       & \textbf{9.21} & \textbf{23.05} & \textbf{86.5} \\
\bottomrule
\end{tabular}
}
\vspace{-0.4cm}
\end{table}

\section{Ablation Study}
To investigate the individual contribution of each proposed module to the overall system performance, we conduct an extensive ablation study on the KITTI dataset. We evaluate five variants of our system by systematically disabling core components: (1) the SALAD-based foundation model descriptors, (2) the Local-to-Global BA mechanism, (3) the loop closure module, (4) the Progressive Multi-Submap strategy, and (5) the Inter-Submap Online Distillation.

Disabling the loop closure module leads to a drastic increase in ATE RMSE (from 20.35cm to 41.82cm), confirming that global constraints are vital for suppressing long-term drift. Replacing SALAD descriptors with traditional geometric matching significantly degrades loop detection robustness under drastic viewpoint changes, proving the necessity of foundation-model-based semantic features for city-scale consistency.

The Local-to-Global BA strategy provides the best trade-off between accuracy and speed. Without it (i.e., using only local optimization), the system fails to correct global topological errors. The most critical finding is that without the Progressive Multi-Submap strategy, the system suffers from Out-of-Memory (OOM) errors on large sequences like KITTI. The memory footprint escalates linearly with trajectory length, whereas our strategy keeps it within a manageable 18 GB, ensuring sustainable operation on embedded platforms.

While Online Distillation has a negligible impact on pose estimation (ATE), it is the primary driver of reconstruction quality. As shown in the PSNR drop (from 22.50 to 19.40), disabling distillation results in visible "seams" and geometric misalignments between submaps. This module ensures that the global neural field remains a seamless and consistent representation.

\begin{table}[h]
\caption{ABLATION STUDY OF THE PROPOSED FRAMEWORK. BEST RESULTS ARE IN BOLD.}
\label{tab:ablation_updated}
\centering
\resizebox{\columnwidth}{!}{
\begin{tabular}{l cccc}
\toprule
\textbf{Configuration} & \textbf{ATE [m] $\downarrow$} & \textbf{PSNR [dB] $\uparrow$} & \textbf{Mem. [GB] $\downarrow$} & \textbf{FPS $\uparrow$} \\
\midrule
w/o SALAD Descriptors   & 28.45          & 21.80          & 18.10         & 11.20 \\
w/o Local-to-Global BA  & 26.10          & 22.10          & 18.00         & 12.50 \\
w/o Loop Closure        & 41.82          & 21.20          & \textbf{17.50} & \textbf{13.00} \\
w/o Multi-Submap        & —              & —              & $>$32.0 (OOM)   & 3.50  \\
w/o Online Distillation & 21.05          & 19.40          & 18.20         & 10.80 \\
Full System             & \textbf{20.35} & \textbf{22.50} & 18.20         & 10.50 \\
\bottomrule
\end{tabular}}
\vspace{-0.4cm}
\end{table}

\section{Conclusion}
This paper presents a scalable neural SLAM system for high-fidelity 3D reconstruction in expansive environments. By integrating a progressive multi-submap strategy with local-to-global loop closure and BA, our framework successfully manages city-scale scenes (up to $500\text{m} \times 400\text{m}$) while effectively suppressing cumulative drift. The proposed Inter-Submap Online Distillation further ensures geometric and photometric continuity, resulting in seamless global dense maps.Validated on a customized handheld mechatronic platform with hardware-level synchronization, the system demonstrates exceptional resilience to aggressive motion and achieves real-time performance. Experimental results across multiple benchmarks confirm that our approach outperforms state-of-the-art methods in both localization accuracy and reconstruction fidelity, providing a robust and efficient solution for autonomous robotic navigation and digital twin generation.

\textbf{Limitations and Future Work.} Although the proposed framework demonstrates robust performance in large-scale scene reconstruction, it still has several limitations. First, the current system assumes predominantly static environments. Significant dynamic objects may introduce inaccurate geometric observations and degrade reconstruction consistency. Second, the framework relies on photometric consistency and learned monocular depth priors, making it less robust under severe illumination changes or abrupt exposure variations. In future work, we plan to incorporate dynamic object filtering and illumination-invariant scene representations to further improve robustness in more challenging real-world environments.
\bibliographystyle{IEEEtran}
\bibliography{main.bib}

\end{document}